\UseRawInputEncoding
\documentclass[letterpaper, 10 pt, journal, twoside]{ieeetran}                                                     

\IEEEoverridecommandlockouts                              % This command is only needed if 
\newcommand{\etal}{\textit{et al}. }
\newcommand{\ie}{\textit{i}.\textit{e}., }
\newcommand{\eg}{\textit{e}.\textit{g}., }

\usepackage{graphicx}
\usepackage{gensymb}
\usepackage{multirow}
\usepackage[font=footnotesize]{caption}
\usepackage{physics,amsmath}
\usepackage{amssymb}
\usepackage{tabularx}
\usepackage{booktabs}
\usepackage{float}
\usepackage[breaklinks,colorlinks,citecolor=black]{hyperref}
\usepackage[noabbrev, capitalize]{cleveref}

\crefformat{section}{\S#2#1#3} 
\crefformat{subsection}{\S#2#1#3}
\crefformat{subsubsection}{\S#2#1#3}
\crefformat{equation}{Eqn.~#2#1#3}
\crefname{figure}{Fig.}{Figs.}
\crefname{table}{Tab.}{Tabs.}

\author{Carter Sifferman$^{1}$, Mohit Gupta$^{1}$, and Michael Gleicher$^{1}$%
\thanks{Manuscript received: May 12, 2025; Revised August 17, 2025; Accepted September 14, 2025.}%Use only for final RAL version
\thanks{This paper was recommended for publication by Editor Pascal Vasseur upon evaluation of the Associate Editor and Reviewers' comments.
This work was supported by NSF Grant 2152163.}
\thanks{$^{1}$All authors are with the Department of Computer Sciences, University of Wisconsin - Madison, USA.
        {\tt\footnotesize [sifferman|mohitg|gleicher]@cs.wisc.edu}}
}

\title{
    Efficient Detection of Objects Near a Robot Manipulator \\
    via Miniature Time-of-Flight Sensors
}

\begin{document}
\maketitle

\begin{abstract}
    We provide a method for detecting and localizing objects near a robot arm using arm-mounted miniature time-of-flight sensors. A key challenge when using arm-mounted sensors is differentiating between the robot itself and external objects in sensor measurements. To address this challenge, we propose a computationally lightweight method which utilizes the raw time-of-flight information captured by many off-the-shelf, low-resolution time-of-flight sensor. We build an empirical model of expected sensor measurements in the presence of the robot alone, and use this model at runtime to detect objects in proximity to the robot. In addition to avoiding robot self-detections in common sensor configurations, the proposed method enables extra flexibility in sensor placement, unlocking configurations which achieve more efficient coverage of a radius around the robot arm. Our method can detect small objects near the arm and localize the position of objects along the length of a robot link to reasonable precision. We evaluate the performance of the method with respect to object type, location, and ambient light level, and identify limiting factors on performance inherent in the measurement principle. The proposed method has potential applications in collision avoidance and in facilitating safe human-robot interaction.

    \smallskip
    \noindent
    Project Page: \href{https://cpsiff.github.io/efficient_detection/}{https://cpsiff.github.io/efficient\_detection/}
\end{abstract}
% \begin{IEEEkeywords}
% Range Sensing, Omnidirectional Vision
% \end{IEEEkeywords}

\section{Introduction}
\begin{figure}
    \centering
    \includegraphics[]{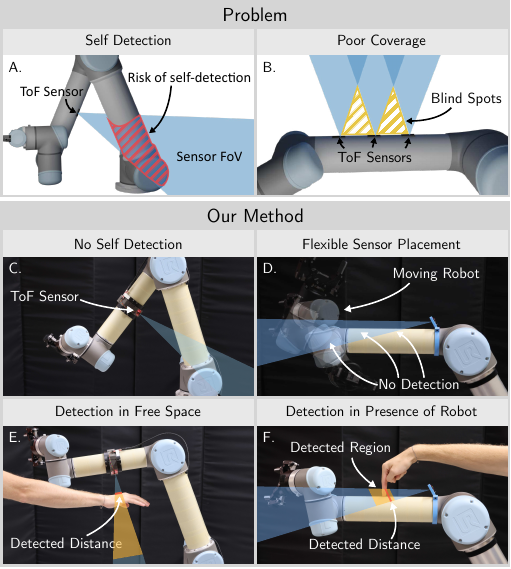}
    \caption{Time-of-flight sensors attached to robot arms are prone to self-detection (A), and typical configurations provide inefficient coverage of the radius near the robot surface (B). Our method enables self-detection free proximity sensing, which enables new sensor configurations that provide more efficient coverage of a radius around the robot surface (C-F).}
    \label{fig:teaser}
    \vspace{-2em}
\end{figure}

\IEEEPARstart{D}{etection} of objects near a robot arm is useful for tasks such as collision avoidance~\cite{Svarny2019Safe, lasota2014toward} or to enable proximity-based human-robot interactions~\cite{escobedo2021contact}. Externally mounted cameras are one way of detecting such objects, but they suffer from occlusion and require the robot to remain in view of the cameras, limiting their practicality when used with mobile manipulators. Therefore, we seek a solution which uses sensors mounted on the robot. Miniature time-of-flight (ToF) sensors~\cite{Callenberg2021CheapSPAD, Sifferman2023unlocking, TMF8820, VL53L8CH} are particularly attractive for for mounting on-robot because of their small size and low power consumption. In order to cover the space around the robot with a small number of sensors, we need the ability to choose efficient sensor configurations, such as placing the sensor peering down the length of an arm segment, as shown in~\cref{fig:teaser}F. However, miniature ToF sensors have very low pixel counts (\eg 3x3) with a wide field-of-view (FoV) per-pixel ($5^\circ$ - $40^\circ$), meaning that in such configurations, the robot is constantly detected, and other objects can only be detected if they are closer than the robot is to the ToF sensor. As illustrated in \cref{fig:filtering-explainer}, simple filtering of pixels which view the robot is ineffective in this case; therefore, to enable such sensor configurations, any approach to object detection must be able to differentiate between \emph{self-detections} (the robot itself) and external object detections \textit{within each pixel}. 

% teaser v4 (one column)

In this work, we provide a method for detecting and localizing objects near a robot arm using miniature ToF sensors. Our method allows for flexibility in sensor placement and avoids self-detections of the robot. We address the key challenge of differentiating robot self-detections from other objects by implicitly modeling the expected appearance, reflectance, and geometry of the robot through sampling of raw ToF measurements. We utilize the raw ToF data captured by commonly used off-the-shelf sensors; at runtime our method finds differences between the measured ToF data and the expected appearance of the robot. Therefore, our method can detect objects even in sensor pixels for which the robot is prominent in view. This enables configurations such as that shown in~\cref{fig:teaser}F, which provide coverage of a small radius around the robot surface using few sensors. Our method also prevents self-detection in more typical outward facing sensor configurations as shown in~\cref{fig:teaser}C.

Our contributions are: 1) a method for detection and localization of objects near a robot arm with a known joint state using a miniature ToF sensor while ignoring robot self-detections; 2) experiments demonstrating that our method is effective at detecting and estimating distance to objects with the configuration shown in~\cref{fig:teaser}F, and experiments investigating the limits and inherent constraints on the performance of our method; and 3) a live demonstration.

\smallskip
\noindent
\textbf{Scope and Limitations.} 
% We don't build a full system, just a one-sensor proof of concept in two configurations
% While it is efficient at run-time, we rely on joint space sampling so it requires lots of pre-processing. Once you've done it the sensor can't be moved without re-doing preprocessing
% Currently not practical for sensor configs where the sensor sees more than 3 DoF
While our method can scale to multiple sensors, in this work we build a prototype which includes one sensor at a time. Our demonstration shows live output of our method, but is not integrated with robot control for \eg collision avoidance, and the sensor frame rate is limited to 3.5 Hz by the data interface of currently available sensors. Our method enables sensor configurations which efficiently cover a small radius around the robot surface, and is computationally efficient at runtime, but requires one-time overnight reference data capture per sensor position. Additionally, our method foregoes the need for any geometric calibration of sensor position, which is required by most alternative methods. 

\section{Related Work}
\subsection{Whole-Robot Proximity Detection}
\label{subsec:whole_body_proximity_detection}

Research on robotic ``sensitive skin'' comprised of touch or proximity sensors dates back to the 1980s~\cite{um1998modularized, lumelsky2000sensitive}. While tactile sensors~\cite{yogeswaran2015new, xia2016multi, liu2022neuro} are useful for collision detection and human-robot interaction, the ability to sense objects before touch occurs (proximity detection) enables a different set of applications including collision avoidance and safety in human-robot interaction. Systems have been proposed for whole-robot proximity detection, including those based on optical ToF~\cite{Tsuji2019, giovinazzo2024cyskin, zhou2023tacsuit, kim_armor_2024}, or other sensing principles~\cite{Fan2021Aurasense, navarro2021proximity}. These works do not address the problem of self-detection directly. Therefore, they are limited to outward facing sensor configurations. Our work demonstrates using flexibility in sensor placement to enable novel configurations that cover regions efficiently.

\subsection{Avoiding Self-Detection}
% \cite{himmelsbach2019single} Weird short not very good paper. But they look at how to avoid detecting the arm itself, but still detecting objects in between two links. They propose to just geometrically figure out what the distance "should" be and compare that to the actual and detect an obstacle when it deviates. They do test on a real arm with a real proximity sensor by locking some DoFs, only 3 dof. Doesn't work well. They blame the material. But also they don't model the whole fov, I say
% \cite{avanzini2014safety} paper trying to optimize the placement of distance sensors on robot. They have a paragraph right before section III where they mention that the sensor might detect the robot or other known things. They use a cad model of the robot and ignore the measurement if it matches what they would expect from the cad model. They build a real prototype
% \cite{wang_robot_2017} they use a kinect external to a robot arm. They're looking at how to remove the signal from the robot arm from the point cloud. Cites many related works

\begin{figure}
    \centering
    \includegraphics[scale=0.8]{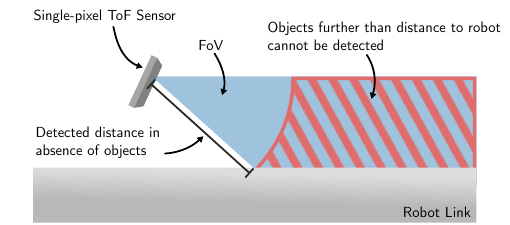}
    \caption{When sensor measurements are treated as single distances per-pixel, objects further than the detected distance to the robot cannot be detected. This leads to significant blind spots, precluding sensor configurations such as the one shown. Our work enables such sensor configurations.}
    \label{fig:filtering-explainer}
    \vspace{-1.5em}
\end{figure}

Self-detection, when the robot itself is detected as an external object, is a challenging problem for robot manipulator perception systems due to the manipulator's dynamic shape during operation. There exists work on avoiding the self-detection problem when using an external depth camera which provides a 3D point cloud, by filtering out points belonging to the robot. Some approaches rely on extrinsic calibration between the camera and the robot, and use a 3D model of the robot to simulate its expected signal in the point cloud~\cite{rakprayoon_kinect-based_2011, sukmanee_obstacle_2012}. Other works do not rely on extrinsic calibration, recognizing and removing the robot from the point cloud directly~\cite{wang_robot_2017}, or using temporal cues along with proprioception~\cite{lyubova_improving_2013, michel_motion-based_2004}. These approaches are a reasonable solution for high resolution point clouds; when the robot points are removed, there is still sufficient information remaining to avoid collisions. However, with a low resolution sensor, filtering point clouds is not effective because the robot may be visible in all or nearly-all pixels. Therefore objects can only be detected when they are closer than the detected distance to the robot in a given pixel, severely limiting the effective detection area for some sensor configurations. A visualization of this limitation is shown in \cref{fig:filtering-explainer}.

There is little prior work on avoiding self-detection with arm-mounted proximity sensors. The works which address the problem directly follow an approach similar to that used for point clouds. Avanzini \etal~\cite{avanzini2014safety} place distance sensors on the links of a robot arm. To avoid self-detections, they use a 3D model of the robot to simulate the expected distance reading if only the robot were present. If the distance reading is less than the simulated reading, an object is detected. Himmelsbach \etal~\cite{himmelsbach2019single} use a similar approach. Such an approach means the sensor will not see objects within its FoV which are further than the simulated distance estimate, again being prone to the problem shown in \cref{fig:filtering-explainer}.

\subsection{Miniature Time-of-Flight Sensors}

Miniature ToF sensors are widely used in robotics due to their small size and low power requirements. Applications have been developed for the sensors mounted on miniature drones~\cite{tsuji2022omnidirectional, muller2023robust, zimmerman2023fully, niculescu2023nanoslam, karam2022microdrone}. Other works place a sparse set of sensors around a robot~\cite{escobedo2021contact, kim_armor_2024} or at a robot wrist~\cite{adamides2019time}, and build applications for collision avoidance. There exist methods which use the sensors to detect the 3DoF pose of a planar surface~\cite{Sifferman2023unlocking}, and methods for calibrating the extrinsic position of a sensor attached to a robot arm~\cite{sifferman2022geometric}. This work builds on previous work which provides a method for detecting geometric deviations on a planar surface~\cite{sifferman2024using}. In contrast to this previous work, the method presented in this paper works for articulated robots and non-planar surfaces, and is able to determine the distance to unknown objects.

There is a body of research which aims to make use of the raw ToF data from low-cost sensors akin to the one used in this work. Callenberg \etal~\cite{Callenberg2021CheapSPAD} demonstrate in-contact material classification and, utilizing additional hardware, high-resolution imaging and non-line-of-sight tracking. Other works look to recover more detailed 3D information from the low-resolution measurements of these sensors. There exist approaches for recovering 3D human pose~\cite{ruget2022pixels2pose}, high resolution depth images~\cite{li2022deltar}, and general 3D reconstruction (from a distributed set of sensors)~\cite{mu20243d}. Miniature ToF sensors have also been used to refine monocular depth estimates~\cite{jungerman2022, nishimura2020disambiguating} and augment RGB SLAM~\cite{tofslam}. Aforementioned work in robotics also takes advantage of raw ToF data~\cite{Sifferman2023unlocking, sifferman2024using}.

\section{Problem Overview}
\begin{figure}
    \centering
    \includegraphics[scale=0.8]{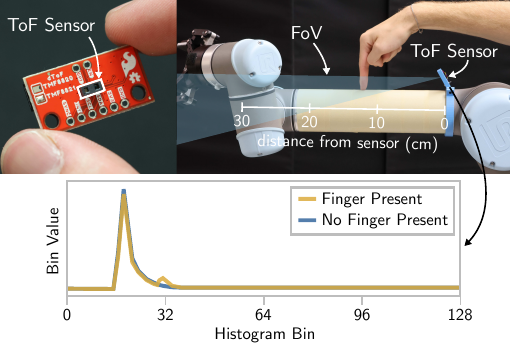}
    \caption{AMS TMF8820 sensor used in this work, and an example histogram captured by the sensor mounted on a robot arm. The large, leftmost peak is due to the surface of the robot link. The presence of the finger causes the appearance of an additional peak.}
    \label{fig:sensor_and_histogram_ex}
    \vspace{-2em}
\end{figure}

\subsection{Background: Direct Time-of-Flight}
The miniature direct ToF sensors that we utilize operate by illuminating a patch of the scene with a pulse of (typically infrared) light and directly measuring the time of travel of the returning light at high (nano- to picosecond) time resolution. The returning light waveform is called the scene \textit{transience}, and the quantized version of that waveform recorded by the sensor is the \textit{transient histogram}~\cite{jungerman2022, gutierrez2022compressive}. The transient histogram is a function of scene geometry and reflectance properties (in addition to sensor and light-source characteristics) integrated over the FoV of the sensor which, for miniature sensors, is typically between 10 and 40 degrees per pixel. Single photon avalanche diodes (SPADs)~\cite{niclass2005design, zappa2007principles} are the most mature and widely available technology enabling direct ToF, and the basis of currently available miniature direct ToF sensors. These sensors are very small ($<$20 mm$^3$), lightweight ($<$1 gram), and power efficient ($<$10 milliwatts per measurement)~\cite{TMF8820, VL53L8CH, VL6180X}.

Typically, miniature ToF sensors use an onboard algorithm to calculate a distance estimate from the transient histogram, which is reported. The goal of this work is to recover the distance to the closest point on unknown geometry while ignoring the robot itself. To accomplish this, our method utilizes the transient histogram directly, rather than on-sensor distance estimates. An example of a transient histogram is shown in \cref{fig:sensor_and_histogram_ex}. Using the transient histogram directly allows us to pick out subtle variations in the measured ToF signal in an individual pixel which are not contained in on-sensor distance estimates alone. In~\cref{subsec:baseline}, we demonstrate that on-sensor distance estimates are not sufficient.

\subsection{Problem Analysis}
\label{subsec:problem_analysis}
In order to estimate the distance to unknown objects in a transient histogram while ignoring the robot itself, we first must identify and remove the signal caused by the robot. In the case of an articulated robot, the geometry of the robot varies with respect to the robot joint state. Our goal is then to create a mapping from robot joint state to a probabilistic model of the expected transient histogram, as it would appear if only the robot were present. This mapping could be achieved in multiple ways. Previous work~\cite{mu20243d, Sifferman2023unlocking} demonstrated an effective forward model for miniature direct ToF sensors, which allows simulation of a histogram measurement given scene geometry and reflectance (\ie surface albedo and specularity). However, in order to accurately simulate sensor measurements of the robot, the spatially varying reflectance properties of the robot would need to be known. Gathering such a measurement requires highly specialized equipment, making a simulation-based approach impractical. Further, previous work has established that detecting objects based on known geometry but unknown reflectance is fundamentally ambiguous under many settings~\cite{sifferman2024using, jungerman2022}.

Rather than modeling the ToF signal of the robot explicitly, we utilize a data-driven approach in which measurements from the sensor are sampled at many robot states with only the robot present. Those measurements are used to create a probabilistic model of the expected transient histogram for any single joint state within the sampled range. Our method is applicable to any direct ToF sensor which reports a transient histogram. We evaluate our method using the AMS TMF8820, shown in~\cref{fig:sensor_and_histogram_ex}.

\section{Method}
Given a $b$-bin transient histogram $\vb{h}_{\text{obs}} \in \mathbb{N}^b$ captured by a miniature ToF sensor attached to an $n$ degree-of-freedom robot with joint state $\vb{q} \in \mathbb{R}^n$, we aim to recover the distance $d$ to the point nearest the sensor on any object in the sensor FoV excluding the robot itself (and any attached accessories, \eg a gripper). In practice, depending on the mounting position of the sensor, the sensor may not be able to see every link of the robot. In this case only degrees-of-freedom which affect sensor readings are included in $\vb{q}$.

\subsection{Histogram Pre-Processing}
\label{subsec:preprocessing}
Transient histograms are affected by ambient light, which manifests as a DC offset in the captured signal~\cite{gupta2019photon}. To avoid falsely detecting changes in ambient light as objects, we pre-process histograms by subtracting the DC offset and normalizing the area under the signal, following the approach of previous work~\cite{sifferman2024using}. For the histogram $\vb{h}$ the DC offset $h_{\text{offset}}$ induced by ambient light is approximated by finding the maximum kernel density on the values of $\vb{h}$, which acts as a robust way of estimating the modal value of $\vb{h}$. The kernel bandwidth $\sigma$ is a tune-able parameter, which can vary by sensor model:
\vspace{-0.5em}
\begin{equation}
    h_{\text{offset}} = \underset{x}{\text{argmax}}\sum_{h_i \in \vb{h}} \mathcal{N}(x; h_i, \sigma)
    \label{eqn:ambient_light_correction}
\end{equation}

The area under the signal is normalized after $h_{\text{offset}}$ is subtracted. The pre-processed histogram $\vb{\tilde{h}}$ is given by:
\begin{equation}
    \label{eqn:normalization}
    \vb{\tilde{h}} = \dfrac{\vb{h} - h_{\text{offset}}}{\|\vb{h} - h_{\text{offset}}\|_1}
\end{equation}

\subsection{Modeling Known Objects}
To detect the distance to unknown objects in the FoV imaged by $\vb{h}_{\text{obs}}$, we rely on a probabilistic model of the per-bin mean $\vb{\mu}_{\vb{q}} \in \mathbb{R}_+^b$ and per-bin variance $\vb{\sigma}_{\vb{q}} \in \mathbb{R}_+^b$ of the expected histogram if only known objects were in the sensor FoV, given the current joint state $\vb{q}$. For reasons explained in~\cref{subsec:problem_analysis}, we approximate $\vb{\mu}_{\vb{q}}$ and $\vb{\sigma}_{\vb{q}}$ by interpolating between real samples over a range of possible $\vb{q}$ rather than an analytical approach.

Our method requires a set of per-bin histogram means $\vb{M}$, variances $\vb{V}$ and corresponding joint angles $\vb{J}$ which sample the robot configuration space. In practice, this dataset can be captured by, \eg grid search or random sampling in configuration space. For each joint position sampled, multiple histograms are captured to generate a good approximation of $\vb{V}$ to capture sensor noise. Details of how we perform this sampling for real-world experiments are given in~\cref{sec:experimental_results}. To approximate $\vb{\mu}_{\vb{q}}$ and $\vb{\sigma}_{\vb{q}}$ from samples in $\vb{M}$ and $\vb{V}$, we perform barycentric interpolation, which requires finding a convex hull of $n+1$ points around $\vb{q}$ in $\vb{J}$. $\vb{\mu}_{\vb{q}}$ and $\vb{\sigma}_{\vb{q}}$ are then interpolated between the corresponding values in $\vb{M}$ and $\vb{V}$. The importance of interpolation and the density of samples needed to achieve good approximation of $\vb{\mu}_{\vb{q}}$ and $\vb{\sigma}_{\vb{q}}$ is investigated in~\cref{subsec:self-detection_rate}.

\subsection{Detecting Distance to Unknown Objects}
\label{subsec:detecting_distance}
Given $\vb{\mu}_{\vb{q}}$ and $\vb{\sigma}_{\vb{q}}$, we calculate the normalized probability vector $\vb{p} \in \mathbb{R}_+^b$ for $\vb{h}_{\text{obs}}$, which encodes the per-bin likelihood that a given bin is in the distribution expected when the robot alone is present, normalized so that for a given bin index $i$, when $h_{\text{obs},i} = \mu_{\vb{q}, i}$ the likelihood is $1$:

\vspace{-0.5em}
\begin{equation}
    \label{eqn:pdf}
    p_i = e^{\dfrac{-(h_{\text{obs},i} - \mu_{\vb{q}, i})^2}{2(\sigma_{\vb{q}, i})^2}}
\end{equation}

To detect objects and estimate their distance, we transform $\vb{p}$ to a binary vector $\vb{g}$ which encodes bins that are likely to contain an unknown object. The threshold for detection $t$ is a hyper-parameter which can be tuned to adjust sensitivity:

\vspace{-0.5em}
\begin{equation}
    g_i = \begin{cases}
        1 & \text{if} \ p_i < t \\
        0 & \text{otherwise}
    \end{cases}
\end{equation}

We then search for segments of the value $1$ in $\vb{g}$ which span $c$ or more contiguous bins, each of which corresponds to one detected object. For each segment, we extract the values of $\vb{h}$ over the corresponding range. We find the peak in these extracted values of $\vb{h}$. The position (bin index) of this peak corresponds to the distance to the detected object. We convert from bin index $i_{\text{peak}}$ to distance using the conversion from bin index to distance for the TMF8820 sensor established by previous work~\cite{Sifferman2023unlocking}: $\text{distance (m)} = 0.01387i_{\text{peak}} - 0.1825$. We empirically observe that this calibration is stable between multiple instances of the same sensor model.

\subsection{TMF8820 Calibration}
\label{subsec:calibration}
We observe that a varying bias is applied to TMF8820 histogram measurements between sensor power cycles. This bias can lead to false positives if the sensor is cycled off after the reference measurements ($\vb{M}, \vb{V}$ and $\vb{J}$) are captured. While we do not know the exact cause of this effect, or if it applies to other ToF sensors, we are able to mitigate it by performing a one-off calibration step every time the sensor is powered on after reference capture. We move the robot to the first reference joint position $\vb{J}_1$ and capture a set of 50 measurements, which we average per-bin and store as $\vb{h}_{\text{ref}}$. We then calculate $\vb{h}_{\text{calib}} = \vb{M}_1 - \vb{h}_{\text{ref}}$. $\vb{h}_{\text{calib}}$ is stored and at query time is added to $\vb{h}$ prior to~\cref{eqn:ambient_light_correction}.

\section{Experimental Results}
\label{sec:experimental_results}

\subsection{Implementation Details}
\label{subsec:implementation_details}
We perform a series of experiments in which a TMF8820 sensor is attached to link two of a Universal Robots UR5 robot arm. The sensor is positioned facing the end effector, as shown in \cref{fig:sensor_and_histogram_ex}. In this position, sensor readings are invariant to movement in the three most proximal joints. Thus, we only sample the 3DoF of the three most distal joints (\ie those comprising the wrist) to capture reference histograms. Each experiment aside from \cref{subsec:ambient_light} relies on the same reference dataset, which is captured over a 3D grid in joint space, in which $\vb{q}_4 \in [-\pi, -\pi/12]$, $\vb{q}_5 \in [-5\pi/6, 5\pi/12]$, $\vb{q}_6 \in [-\pi/2, 5\pi/12]$. Joint positions are sampled in $\pi/12$ radian increments, for a total of 2,304 joint positions. It takes $\sim10$ hours to programmatically capture 50 measurements per joint position. The brushed aluminum surface of the UR5 robot is highly specular. While our data-driven approach models the effect of the specular surface when the robot alone is present, when other objects are present outside of the sensor field-of-view the surface acts like a mirror. This leads to false detections when objects outside of the sensor FoV are detected via three-bounce paths. We cover the metallic surfaces of the robot in masking tape to minimize this effect, and further investigate in \cref{subsec:NLOS_objects}

% 23040 measurements @ 3.5 FPS = 6582 seconds = 109 minutes (lower bound)
% 115200 measurements @ 3.5 FPS = 32914 seconds = 548 minutes = 9.14 hours

We capture data from an AMS TMF8820 sensor connected to an Arduino microcontroller, using the microcontroller code provided by prior works~\cite{Sifferman2023unlocking, mu20243d, sifferman2024using} to extract both distance and histogram measurements. The TMF8820 reports 9 histograms over 9 non-overlapping zones, for a total FoV of $30^\circ$ diagonally. For simplicity and to limit the FoV to avoid unwanted detections, we utilize only one zone of the sensor, yielding a $10^\circ$ diagonal FoV. Sensor frame rate is limited by the speed at which the I$^2$C interface, which is not designed for histogram data capture, can transmit histograms, so we modify the microcontroller code to only report bins 1-80 to increase frame rate. This means the maximum range of the sensor as configured is $\sim90$cm. The sensor reports measurements at 3.5 FPS. The execution of our algorithm takes 0.35 ms (2803 FPS) on a mid-range laptop CPU (Intel i5 1340P), and the runtime scales linearly with the number of sensors used. In our testing, interference between multiple TMF8820 sensors is minimal, making them well-suited to future systems with many sensors.

Unless otherwise stated, we set the probability threshold $t = 0.001$, and minimum segment size $c = 4$. These values were manually tuned to create a reasonably low false positive rate for the main experiments. We investigate the effect of changing these parameters in~\cref{subsec:parameter_tuning_and_ablation}. A peak in bin $\sim$14 from the TMF8820 corresponds to an object at distance zero; therefore we trim the histogram $\vb{\tilde{h}}$ to bin range $(15, 80)$ before applying~\cref{eqn:pdf}. We set $\sigma$ in~\cref{eqn:ambient_light_correction} to 5, following previous work~\cite{sifferman2024using}. In each experiment, we only consider the \textit{closest} detection. All captures are  in a windowless room with fluorescent lights ($\sim500$ lux).

\subsection{Self-Detection Rate}
\label{subsec:self-detection_rate}

\begin{figure}
    \centering
    \includegraphics[scale=0.9]{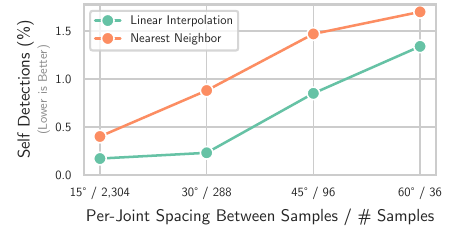}
    \caption{Effect of joint-space sampling density on self-detection (false positive) rate. Linear interpolation of background histograms between nearby joint states outperforms nearest neighbor interpolation (lower is better). 3 DoF of robot wrist joints are sampled.}
    \label{fig:self_detection_plot}
    \vspace{-1em}
\end{figure}

We perform an experiment to understand the effect of joint-space sampling density on the rate at which the robot is falsely detected. The results of this experiment are highly dependent on robot geometry and sensor position. The experiment serves to provide a rough approximation of performance in general, and provides context for other experiments, which use the same robot and the same sensor position.

We capture a dataset of ToF measurements from 1000 uniformly sampled random joint positions within the joint range of the reference dataset, with only the robot present. The self-detection rate (false positive rate) is the rate at which detections occur in this dataset. To investigate the effect that sampling density has on self-detection rate, we sub-sample the reference dataset by a factor of 2, 3, and 4 per dimension, creating a coarser grid of samples on which the model is built, and plot the effect on self-detection rate in~\cref{fig:self_detection_plot}. Self-detection rate increases as the density of joint-space samples decreases, and linear interpolation leads to a lower self-detection rate than nearest neighbor interpolation. Self-detection rate levels off at high sampling densities; we hypothesize that this is because in rare cases measurement noise leads to self-detections, and measurement noise is constant regardless of sample density. A coarser sampling of every $30\degree$ achieves similar performance to $15\degree$ while requiring an order of magnitude fewer samples. Coupled with future sensors with a higher frame rate, this means that reference data capture could be made orders of magnitudes faster for little performance penalty.

\subsection{True Positive Rate}
\label{subsec:true_positive_rate}

\begin{figure}
    \centering
    \includegraphics[]{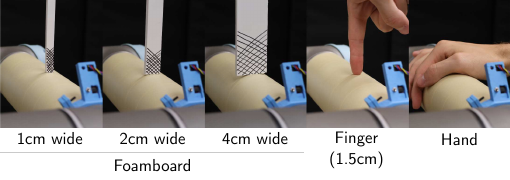}
    \caption{Objects used in experiments, shown as placed on the robot for data capture. Pieces of foamboard have a hatching pattern applied to provide visual features for ground-truth depth-from-stereo camera.}
    \label{fig:objects}
    \vspace{-2em}
\end{figure}

\begin{figure}
    \centering
    \includegraphics[scale=0.9]{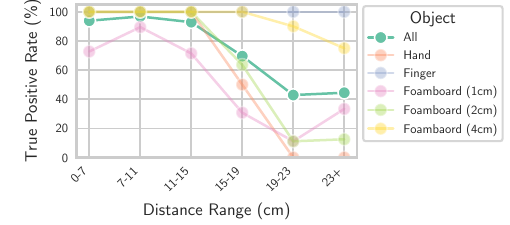}
    \caption{True positive rate as a factor of the distance from the sensor to the object, broken down by object type. Note that due to random object placement, each data point may not represent the same number of samples.}
    \label{fig:true_positive_rate}
    \vspace{-2em}
\end{figure}

To evaluate the true positive rate of our method (\ie the rate at which an object is detected when one is present), we capture a dataset of ToF measurements in which objects are touching or nearly touching the robot arm. Between each measurement, the robot is moved to a random uniformly sampled joint state within the bounds of the reference dataset, and moved the object to a random distance from the sensor along the robot arm ($1$-$28\text{cm}$ from the sensor); random distances are used to make data capture faster, ultimately allowing a larger dataset. We utilize five objects: a human pointer finger, a human hand, and long pieces of white foamboard cut to 1cm, 2cm, and 4cm in width. The objects are shown in~\cref{fig:objects}. The finger and hand were captured touching the robot arm, while each piece of foamboard was captured at 0cm, 1cm, and 2cm proximity from the arm itself. In total, the dataset contains 275 captures, each with varying conditions (object, distance to sensor, and proximity to arm).

We achieve a true positive rate of $78.9\%$; this is broken down by object and distance to the sensor in~\cref{fig:true_positive_rate}. The 4cm wide foamboard and pointer finger are the easiest to detect at all distances, while the narrower foamboard and hand are the most difficult. The hand, 1cm, and 2cm-wide foamboard are rarely recognized beyond 19cm from the sensor. We hypothesize that the difference between objects is due to a difference in their cross-sectional area and geometric deviation from the robot. While the hand is a large object, its cross sectional area is small from the point-of-view of the sensor when the hand is resting flat on the robot, and the hand does not extend far above the robot surface, leading to a relatively small change in the histogram. Object albedo might also play a factor; an object which is much brighter or darker than the robot will cause a larger change in the measured histogram than one with the same albedo. Lastly, non-line-of-sight effects (see \cref{subsec:NLOS_objects}) caused by the rest presence of the wrist and rest of the arm above the finger may make it easier to detect than the narrow foamboard. Future work should aim to isolate and negate the cause of the performance gap between objects.

\subsection{Distance Estimation}
\label{subsec:distance_estimation}

\begin{figure}
    \centering
    \includegraphics[]{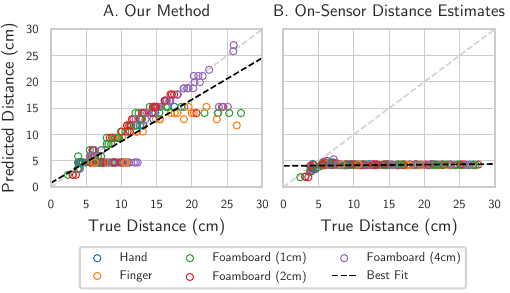}
    \caption{A. Actual distance vs. distance predicted by our method. We achieve an average distance error of 2.08cm. B. Actual distance vs. distance predicted by a baseline method which utilizes on-sensor distance estimates. The baseline method never estimates a distance further than 5cm due to the limitation illustrated in \cref{fig:filtering-explainer}.}
    \label{fig:distance_estimation_scatter}
    \vspace{-1em}
\end{figure}

We use the same dataset as in the previous subsection~(\cref{subsec:true_positive_rate}) to evaluate the accuracy of our distance estimate. Ground truth distance labels are captured via an Intel Realsense D405 depth-from-stereo camera positioned next to the ToF sensor. The closest point on each object is labeled, and the depth estimate extracted from the D405 depth image. Objects closer than the minimum depth range of the camera are labeled by overlaying an image of the robot arm with ruler marks onto the captured RGB image of the object.

\cref{fig:distance_estimation_scatter}A compares the distance estimate from our method to the actual distance. We achieve an average absolute error of 2.08cm. Our method under-estimates the distance to objects in some cases. One case is when a nearby object fills a large portion of the FoV, completely changing the shape of the histogram. When this happens, it is difficult to align the observed histogram to the reference to localize the deviation. Our method also sometimes under-estimates the distance to far away objects. We hypothesize that this could be due to a low signal-to-noise ratio, and/or the presence of the dynamic robot wrist links at those distances.

\subsection{On-Sensor Distance Estimation}
\label{subsec:baseline}
We compare the distance estimation results achieved by on-sensor distance estimates to our method. The TMF8820 reports up to two distance estimates per-zone, corresponding to up to two objects in the FoV. For each sample we choose the distance estimate of the two which achieves the lowest absolute error from the ground truth to demonstrate the best case for on-sensor distance estimates. In~\cref{fig:distance_estimation_scatter}B, we show the distance estimated by this method compared to the true distance. The distance estimates never exceed 5cm, roughly the distance to the nearest point on the robot. This experiment demonstrates the limitation illustrated in \cref{fig:filtering-explainer} and makes it clear that the results achieved by our method cannot be achieved using on-sensor distance estimates alone.

\subsection{Sensor Field-of-View}
\label{subsec:sensor_fov}

\begin{figure}
    \centering
    \includegraphics[width=\linewidth]{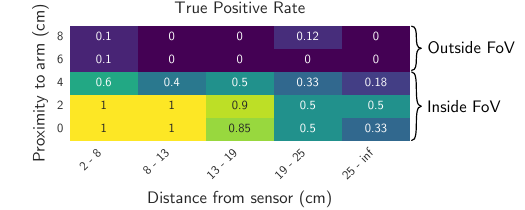}
    \caption{Detection rate as a factor of object distance from the sensor and proximity to the arm. We see that objects beyond 4cm proximity to the arm are rarely detected, and that the proximity of detection is consistent regardless of distance from the sensor. This well-defined field-of-view is desirable for downstream applications.}
    \label{fig:fov_heatmap}
    \vspace{-1em}
\end{figure}

We characterize the FoV of the sensor by placing the 4cm and 2cm foamboard at varying distance from the sensor and proximity to the arm and plot the TPR per object position in~\cref{fig:fov_heatmap}. The sensor is 4cm from the robot surface, with the top edge of the FoV aligned with the surface. Accordingly, we see that detection is less likely at 4cm from the surface and much less likely at 6cm+. The maximum detection proximity is consistent across all distances from the sensor, making the sensor configuration effective for detecting objects that come within 4cm of the robot surface.

\subsection{Non-Line-of-Sight Objects}
\label{subsec:NLOS_objects}

\begin{figure}
    \centering
    \includegraphics[scale=0.8]{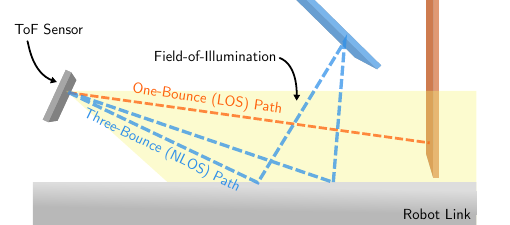}
    \caption{There is an ambiguity between distant line-of-sight (LOS) objects and nearer non-line-of-sight (NLOS) objects outside of the sensor field-of-illumination. This limits the performance of our method on distant objects. If the detection threshold is lowered to determine distance LOS objects, nearer NLOS objects will also be detected.}
    \label{fig:nlos_explainer}
    \vspace{-2em}
\end{figure}

Direct ToF sensors are subject to non-line-of-sight (NLOS) effects, which occur when photons bounce multiple times before returning to the sensor, as illustrated by the blue path in \cref{fig:nlos_explainer}. As previously noted, this effect is very noticeable with the brushed aluminum robot surface. We cover the arm with masking tape to lower this effect for our experiments, but it is still present. In this subsection, we investigate the prevalence of NLOS effects in our setting.

\begin{figure}
    \centering
    \includegraphics[]{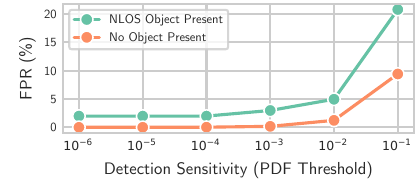}
    \caption{When there is a large object outside of the sensor field-of-view, false positives are more likely due to three-bounce paths from the object.}
    \label{fig:nlos_vs_no_object_fpr}
    \vspace{-0.5em}
\end{figure}

We investigate NLOS effects over the previously captured dataset (\cref{subsec:sensor_fov}). We treat objects outside of the direct FoI of the sensor (\ie, 6cm and 8cm proximity to the arm; the top two rows in \cref{fig:fov_heatmap}) as NLOS objects. We compare false positive rate between this set of objects and the case where only the robot is present. As shown in \cref{fig:nlos_vs_no_object_fpr}, false positives are much more likely when an NLOS object is present. The NLOS effect limits the performance of our system when the detection threshold is limited to not detect NLOS objects; increasing the sensitivity to detect more distant objects comes at the expense of increased detection of NLOS objects. On the other hand, for some applications, detection of NLOS objects may be welcome. In such cases the sensitivity can be increased to detect more distant line-of-sight objects with little increase in false positives in the absence of any objects.

\subsection{Parameter Tuning and Ablation Study}
\label{subsec:parameter_tuning_and_ablation}

\begin{figure}
    \centering
    \includegraphics[]{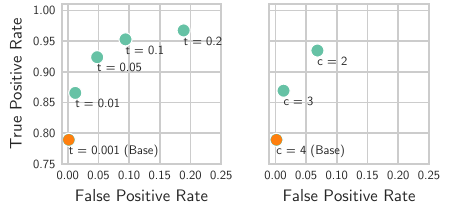}
    \caption{Changing method parameters ($t$ or $c$ described in \cref{subsec:detecting_distance}) leads to different operating points on the ROC curve.}
    \label{fig:ROC_tradeoff_plot}
    \vspace{-1.5em}
\end{figure}

\begin{table}[]
    \centering
    \begin{tabular}{lrr}
         Condition & TPR ($\uparrow$) & FPR ($\downarrow$) \\
         \cmidrule(){1-3}
         Base & 0.789 & 0.002 \\
         No Preprocessing & 0.785 & 0.002 \\
         No Calibration & 0.996 & 0.917 \\
         No Bin Trimming & 0.938 & 0.706 \\
    \end{tabular}
    \caption{Ablation study. Calibration and bin trimming are necessary to avoid high false positives. Preprocessing has no effect as the test dataset does not contain changes in ambient light.}
    \label{tab:ablation_study}
    % \vspace{-2em}
\end{table}

Tuning the parameters $t$ and $c$ (see~\cref{subsec:detecting_distance}) allows for a tradeoff between false positive rate and true positive rate. We use the dataset from~\cref{subsec:self-detection_rate} to test false positive rate and that from~\cref{subsec:true_positive_rate} (containing objects 0cm, 1cm, and 2cm from the robot surface) to test true positive rate. The operating points achieved by varying $t$ and $c$ are shown in~\cref{fig:ROC_tradeoff_plot}. Varying these parameters allows for tuning of the FPR/TPR trade-off for different scenarios. However, as shown in \cref{fig:nlos_vs_no_object_fpr}, increasing the sensitivity of the method also increases the detection rate of non-line-of-sight objects.

We ablate components of our method and show the effect on true positive rate and false positive rate in~\cref{tab:ablation_study}. We find that calibration~(\cref{subsec:calibration}), and bin trimming are necessary to prevent a high false positive rate. Normalization~(\cref{eqn:ambient_light_correction}) has no significant effect on the results  because the test dataset does not exhibit strong changes in ambient light. 

\subsection{Performance Under Varying Ambient Light}
\label{subsec:ambient_light}
Ambient light affects the histogram, potentially leading to a false positive if ambient light level changes after reference capture. To evaluate our method under such changes, we capture new reference captures at the same joint locations as previous experiments. For each of four ambient light levels, 100 measurements of the robot are captured at random joint positions with only the robot present. We calculate the false positive rate of our method at each of these four light levels. The results are shown in \cref{tab:ambient_light}. Moderate changes in ambient light from the reference capture (which was taken under the ``Dim LED'' lighting) do not lead to an increase in false positive rate. The bright, IR-heavy halogen light source leads to a sharp increase in false positive rate. This means our method is impractical when ambient light changes significantly (\eg moving from indoors to sunlight). Additionally, we observe that the pre-processing steps described in \cref{subsec:preprocessing} are somewhat effective at improving performance under high ambient light. It is possible that more sophisticated methods for reversing the effect of ambient light \cite{gupta2019photon} could improve the performance of our method in these scenarios, but they require an accurate model of sensor hardware which is not available with current commercially available sensors.

\begin{table}[]
    \centering
    \begin{tabular}{lrrr}
         & & \multicolumn{2}{c}{FPR ($\downarrow$)} \\
         Lighting & Lux &  Base Method & No Pre-Processing \\
         \cmidrule(){1-4}
         Dark & $<0.1$ & 0.0099 & 0.0198 \\
         Dim LEDs$^*$ & 100 & 0.0198 & 0.0198 \\
         Flourescent Lights & 500 & 0.0099 & 0.1089 \\
         Halogen Lights & 1000 & 0.1782 & 1.0000 \\
    \end{tabular}
    \caption{Evaluation of the false positive rate of our method under varying levels of ambient light. Poor performance is observed under bright halogen lights, and performance is worse with no pre-processing (\cref{eqn:normalization}). $^*$~Dim LED lights match the lighting used during reference capture.}
    \label{tab:ambient_light}
    \vspace{-2em}
\end{table}

\section{Demonstration}
\label{sec:demo}
% \begin{figure}
%     \centering
%     \includegraphics[]{media/demo_v2.pdf}
%     \caption{Demonstration of our method in use with a sensor position common in prior work. A. The robot itself is not detected. B. An unknown object (human hand) is detected.}
%     \label{fig:demo}
% \end{figure}
We demonstrate our method on a UR5 robot arm. As objects approach the arm, they are detected and their position shown overlaid on an image of the robot as a yellow region corresponding to the range of the detected segments, and a red line for the detected distance, shown in~\cref{fig:teaser}E and ~\cref{fig:teaser}F. In addition to the configuration used for experiments~(\cref{sec:experimental_results}), we demonstrate an additional configuration in which the sensor is positioned orthogonal to the robot surface. The only change made to accommodate this configuration is that joints 2 and 3 of the robot are sampled for reference measurements rather than joints 4, 5, and 6. A live demonstration of both sensor configurations is provided in the supplementary video. This demonstration shows that our method is effective under multiple sensor configurations with minimal adjustments.

\section{Conclusion and Future Work}
% more extensive validation in other sensor configurations
% figuring out what the best sensor configurations are
% better hardware design
% building robot systems which utilize this technique, which will require hardware design, kinematics, and controls
% adaptive sampling or pre-planned varied sample density based on robot geometry
% sensors with better frame rates
% understand calibration

This work demonstrates that it is possible to use raw ToF measurements to extract information about objects near a robot arm that would be impossible to obtain from distance estimates alone. We see poor performance on small objects when they are placed beyond $\sim15$cm from the sensor, limited partially by non-line-of-sight effects. Future work should investigate fusing sensor measurements to resolve the ambiguity between line-of-sight and non-line-of-sight objects. Our method could also be made more practical. We sample joint space uniformly to capture reference data, which is inefficient; future work should explore adaptive sampling or vary sampling density based on robot geometry. Creating a robot system which utilizes our method will require improved sensing hardware design (\ie PCBs and integrated wiring) to decrease footprint and increase frame rate, in addition to integration with kinematics and control algorithms, akin to \cite{escobedo2021contact, kim_armor_2024, rakita_collisionik_2021}, to enable human-robot interaction and collision avoidance. A full system could additionally leverage temporal filtering to improve detection stability and performance. Further work should also further investigate performance as a factor of the sensor placement and the geometry and reflectance properties of detected objects. We believe the method presented in this work is a step towards whole-body proximity sensing with minimal hardware sensing cost.


\bibliography{references}

\begin{thebibliography}{10}
\providecommand{\url}[1]{#1}
\csname url@rmstyle\endcsname
\providecommand{\newblock}{\relax}
\providecommand{\bibinfo}[2]{#2}
\providecommand\BIBentrySTDinterwordspacing{\spaceskip=0pt\relax}
\providecommand\BIBentryALTinterwordstretchfactor{4}
\providecommand\BIBentryALTinterwordspacing{\spaceskip=\fontdimen2\font plus
\BIBentryALTinterwordstretchfactor\fontdimen3\font minus \fontdimen4\font\relax}
\providecommand\BIBforeignlanguage[2]{{%
\expandafter\ifx\csname l@#1\endcsname\relax
\typeout{** WARNING: IEEEtran.bst: No hyphenation pattern has been}%
\typeout{** loaded for the language `#1'. Using the pattern for}%
\typeout{** the default language instead.}%
\else
\language=\csname l@#1\endcsname
\fi
#2}}

\bibitem{Svarny2019Safe}
P.~Svarny, M.~Tesar, J.~K. Behrens, and M.~Hoffmann, ``Safe physical hri: Toward a unified treatment of speed and separation monitoring together with power and force limiting,'' in \emph{International Conference on Intelligent Robots and Systems}, 2019, pp. 7580--7587.

\bibitem{lasota2014toward}
P.~A. Lasota, G.~F. Rossano, and J.~A. Shah, ``Toward safe close-proximity human-robot interaction with standard industrial robots,'' in \emph{International Conference on Automation Science and Engineering}.\hskip 1em plus 0.5em minus 0.4em\relax IEEE, 2014, pp. 339--344.

\bibitem{escobedo2021contact}
C.~Escobedo, M.~Strong, M.~West, A.~Aramburu, and A.~Roncone, ``Contact anticipation for physical human--robot interaction with robotic manipulators using onboard proximity sensors,'' in \emph{International Conference on Intelligent Robots and Systems}.\hskip 1em plus 0.5em minus 0.4em\relax IEEE, 2021, pp. 7255--7262.

\bibitem{Callenberg2021CheapSPAD}
C.~Callenberg, Z.~Shi, F.~Heide, and M.~B. Hullin, ``Low-cost spad sensing for non-line-of-sight tracking, material classification and depth imaging,'' \emph{ACM Transactions on Graphics}, vol.~40, no.~4, 2021.

\bibitem{Sifferman2023unlocking}
C.~Sifferman, Y.~Wang, M.~Gupta, and M.~Gleicher, ``Unlocking the performance of proximity sensors by utilizing transient histograms,'' \emph{Robotics and Automation Letters}, vol.~8, no.~10, pp. 6843--6850, 2023.

\bibitem{TMF8820}
{AMS OSRAM AG}, \emph{\BIBforeignlanguage{English}{TMF882X Datasheet}}, AMS OSRAM AG.

\bibitem{VL53L8CH}
{ST Microelectronics}, \emph{\BIBforeignlanguage{English}{VL53L8CH Datasheet}}, ST Microelectronics.

\bibitem{um1998modularized}
D.~Um, B.~Stankovic, K.~Giles, T.~Hammond, and V.~Lumelsky, ``A modularized sensitive skin for motion planning in uncertain environments,'' in \emph{International Conference on Robotics and Automation}, vol.~1.\hskip 1em plus 0.5em minus 0.4em\relax IEEE, 1998, pp. 7--12.

\bibitem{lumelsky2000sensitive}
V.~Lumelsky, M.~S. Shur, S.~Wagner, and M.~Ding, \emph{Sensitive skin}.\hskip 1em plus 0.5em minus 0.4em\relax World Scientific, 2000, vol.~18.

\bibitem{yogeswaran2015new}
N.~Yogeswaran, W.~Dang, W.~T. Navaraj, D.~Shakthivel, S.~Khan, E.~O. Polat, S.~Gupta, H.~Heidari, M.~Kaboli, L.~Lorenzelli, \emph{et~al.}, ``New materials and advances in making electronic skin for interactive robots,'' \emph{Advanced Robotics}, vol.~29, no.~21, pp. 1359--1373, 2015.

\bibitem{xia2016multi}
F.~Xia, B.~Bahreyni, and F.~Campi, ``Multi-functional capacitive proximity sensing system for industrial safety applications,'' in \emph{Sensors}.\hskip 1em plus 0.5em minus 0.4em\relax IEEE, 2016, pp. 1--3.

\bibitem{liu2022neuro}
F.~Liu, S.~Deswal, A.~Christou, Y.~Sandamirskaya, M.~Kaboli, and R.~Dahiya, ``Neuro-inspired electronic skin for robots,'' \emph{Science robotics}, vol.~7, no.~67, p. eabl7344, 2022.

\bibitem{Tsuji2019}
S.~{Tsuji} and T.~{Kohama}, ``Proximity skin sensor using time-of-flight sensor for human collaborative robot,'' \emph{Sensors}, vol.~19, no.~14, pp. 5859--5864, 2019.

\bibitem{giovinazzo2024cyskin}
F.~Giovinazzo, F.~Grella, M.~Sartore, M.~Adami, R.~Galletti, and G.~Cannata, ``From cyskin to proxyskin: Design, implementation and testing of a multi-modal robotic skin for human--robot interaction,'' \emph{Sensors}, vol.~24, no.~4, p. 1334, 2024.

\bibitem{zhou2023tacsuit}
Y.~Zhou, J.~Zhao, P.~Lu, Z.~Wang, and B.~He, ``Tacsuit: A wearable large-area, bioinspired multi-modal tactile skin for collaborative robots,'' \emph{Transactions on Industrial Electronics}, 2023.

\bibitem{kim_armor_2024}
D.~Kim, M.~Srouji, C.~Chen, and J.~Zhang, ``\BIBforeignlanguage{en}{{ARMOR}: {Egocentric} {Perception} for {Humanoid} {Robot} {Collision} {Avoidance} and {Motion} {Planning}},'' Nov. 2024, arXiv:2412.00396 [cs].

\bibitem{Fan2021Aurasense}
X.~Fan, R.~Simmons-Edler, D.~Lee, L.~Jackel, R.~Howard, and D.~Lee, ``Aurasense: Robot collision avoidance by full surface proximity detection,'' in \emph{International Conference on Intelligent Robots and Systems}, 2021, pp. 1763--1770.

\bibitem{navarro2021proximity}
S.~E. Navarro, S.~M{\"u}hlbacher-Karrer, H.~Alagi, H.~Zangl, K.~Koyama, B.~Hein, C.~Duriez, and J.~R. Smith, ``Proximity perception in human-centered robotics: A survey on sensing systems and applications,'' \emph{Transactions on Robotics}, vol.~38, no.~3, pp. 1599--1620, 2021.

\bibitem{rakprayoon_kinect-based_2011}
P.~Rakprayoon, M.~Ruchanurucks, and A.~Coundoul, ``Kinect-based obstacle detection for manipulator,'' in \emph{{International} {Symposium} on {System} {Integration}}, Dec. 2011, pp. 68--73.

\bibitem{sukmanee_obstacle_2012}
W.~Sukmanee, M.~Ruchanurucks, and P.~Rakprayoon, ``Obstacle modeling for manipulator using iterative least square ({ILS}) and iterative closest point ({ICP}) base on {Kinect},'' in \emph{{International} {Conference} on {Robotics} and {Biomimetics}}, Dec. 2012, pp. 672--676.

\bibitem{wang_robot_2017}
X.~Wang, C.~Yang, Z.~Ju, H.~Ma, and M.~Fu, ``\BIBforeignlanguage{en}{Robot manipulator self-identification for surrounding obstacle detection},'' \emph{\BIBforeignlanguage{en}{Multimedia Tools and Applications}}, vol.~76, no.~5, pp. 6495--6520, Mar. 2017.

\bibitem{lyubova_improving_2013}
N.~Lyubova, D.~Filliat, and S.~Ivaldi, ``Improving object learning through manipulation and robot self-identification,'' in \emph{{International} {Conference} on {Robotics} and {Biomimetics}}, Dec. 2013, pp. 1365--1370.

\bibitem{michel_motion-based_2004}
P.~Michel, K.~Gold, and B.~Scassellati, ``Motion-based robotic self-recognition,'' in \emph{International Conference on Robotics and Systems}, vol.~3, Sept. 2004, pp. 2763--2768 vol.3.

\bibitem{avanzini2014safety}
G.~Buizza~Avanzini, N.~M. Ceriani, A.~M. Zanchettin, P.~Rocco, and L.~Bascetta, ``Safety {Control} of {Industrial} {Robots} {Based} on a {Distributed} {Distance} {Sensor},'' \emph{Transactions on Control Systems Technology}, vol.~22, no.~6, pp. 2127--2140, Nov. 2014.

\bibitem{himmelsbach2019single}
U.~B. Himmelsbach, T.~M. Wendt, N.~Hangst, and P.~Gawron, ``Single pixel time-of-flight sensors for object detection and self-detection in three-sectional single-arm robot manipulators,'' in \emph{International Conference on Robotic Computing}.\hskip 1em plus 0.5em minus 0.4em\relax IEEE, 2019, pp. 250--253.

\bibitem{tsuji2022omnidirectional}
S.~Tsuji and T.~Kohama, ``Omnidirectional proximity sensor system for drones using optical time-of-flight sensors,'' \emph{Transactions on Electrical and Electronic Engineering}, vol.~17, no.~1, pp. 19--25, 2022.

\bibitem{muller2023robust}
H.~M{\"u}ller, V.~Niculescu, T.~Polonelli, M.~Magno, and L.~Benini, ``Robust and efficient depth-based obstacle avoidance for autonomous miniaturized uavs,'' \emph{Transactions on Robotics}, 2023.

\bibitem{zimmerman2023fully}
N.~Zimmerman, H.~M{\"u}ller, M.~Magno, and L.~Benini, ``Fully onboard low-power localization with semantic sensor fusion on a nano-uav using floor plans,'' \emph{arXiv preprint arXiv:2310.12536}, 2023.

\bibitem{niculescu2023nanoslam}
V.~Niculescu, T.~Polonelli, M.~Magno, and L.~Benini, ``Nanoslam: Enabling fully onboard slam for tiny robots,'' \emph{Internet of Things Journal}, 2023.

\bibitem{karam2022microdrone}
S.~Karam, F.~Nex, B.~T. Chidura, and N.~Kerle, ``Microdrone-based indoor mapping with graph slam,'' \emph{Drones}, vol.~6, no.~11, p. 352, 2022.

\bibitem{adamides2019time}
O.~A. Adamides, A.~S. Modur, S.~Kumar, and F.~Sahin, ``A time-of-flight on-robot proximity sensing system to achieve human detection for collaborative robots,'' in \emph{International Conference on Automation Science and Engineering}.\hskip 1em plus 0.5em minus 0.4em\relax IEEE, 2019, pp. 1230--1236.

\bibitem{sifferman2022geometric}
C.~Sifferman, D.~Mehrotra, M.~Gupta, and M.~Gleicher, ``Geometric calibration of single-pixel distance sensors,'' \emph{Robotics and Automation Letters}, vol.~7, no.~3, pp. 6598--6605, 2022.

\bibitem{sifferman2024using}
C.~Sifferman, W.~Sun, M.~Gupta, and M.~Gleicher, ``Using a distance sensor to detect deviations in a planar surface,'' \emph{Robotics and Automation Letters}, vol.~9, no.~10, pp. 8515--8522, 2024.

\bibitem{ruget2022pixels2pose}
A.~Ruget, M.~Tyler, G.~Mora~Mart{\'\i}n, S.~Scholes, F.~Zhu, I.~Gyongy, B.~Hearn, S.~McLaughlin, A.~Halimi, and J.~Leach, ``Pixels2pose: Super-resolution time-of-flight imaging for 3d pose estimation,'' \emph{Science Advances}, vol.~8, no.~48, p. eade0123, 2022.

\bibitem{li2022deltar}
Y.~Li, X.~Liu, W.~Dong, H.~Zhou, H.~Bao, G.~Zhang, Y.~Zhang, and Z.~Cui, ``Deltar: Depth estimation from a light-weight tof sensor and rgb image,'' in \emph{European Conference on Computer Vision}.\hskip 1em plus 0.5em minus 0.4em\relax Springer, 2022, pp. 619--636.

\bibitem{mu20243d}
F.~Mu, C.~Sifferman, S.~Jungerman, Y.~Li, M.~Han, M.~Gleicher, M.~Gupta, and Y.~Li, ``Towards 3d vision with low-cost single-photon cameras,'' in \emph{Computer Vision and Pattern Recognition}, 2024, pp. 5302--5311.

\bibitem{jungerman2022}
S.~Jungerman, A.~Ingle, Y.~Li, and M.~Gupta, ``3d scene inference from transient histograms,'' in \emph{European Conference on Computer Vision}.\hskip 1em plus 0.5em minus 0.4em\relax Springer, 2022, pp. 401--417.

\bibitem{nishimura2020disambiguating}
M.~Nishimura, D.~B. Lindell, C.~Metzler, and G.~Wetzstein, ``Disambiguating monocular depth estimation with a single transient,'' in \emph{European Conference on Computer Vision}.\hskip 1em plus 0.5em minus 0.4em\relax Springer, 2020, pp. 139--155.

\bibitem{tofslam}
L.~Xinyang, L.~Yijin, T.~Yanbin, B.~Hujun, Z.~Guofeng, Z.~Yinda, and C.~Zhaopeng, ``Multi-modal neural radiance field for monocular dense slam with a light-weight tof sensor,'' in \emph{International Conference on Computer Vision}, 2023.

\bibitem{gutierrez2022compressive}
F.~Gutierrez-Barragan, A.~Ingle, T.~Seets, M.~Gupta, and A.~Velten, ``Compressive single-photon 3d cameras,'' in \emph{Computer Vision and Pattern Recognition}, 2022, pp. 17\,854--17\,864.

\bibitem{niclass2005design}
C.~Niclass, A.~Rochas, P.-A. Besse, and E.~Charbon, ``Design and characterization of a cmos 3-d image sensor based on single photon avalanche diodes,'' \emph{Journal of Solid-State Circuits}, vol.~40, no.~9, pp. 1847--1854, 2005.

\bibitem{zappa2007principles}
F.~Zappa, S.~Tisa, A.~Tosi, and S.~Cova, ``Principles and features of single-photon avalanche diode arrays,'' \emph{Sensors and Actuators A: Physical}, vol. 140, no.~1, pp. 103--112, 2007.

\bibitem{VL6180X}
{ST Microelectronics}, \emph{\BIBforeignlanguage{English}{VL6180X Proximity and Ambient Light Sensing Module Datasheet}}, ST Microelectronics.

\bibitem{gupta2019photon}
A.~Gupta, A.~Ingle, A.~Velten, and M.~Gupta, ``Photon-flooded single-photon 3d cameras,'' in \emph{Computer Vision and Pattern Recognition}, 2019, pp. 6770--6779.

\bibitem{rakita_collisionik_2021}
D.~Rakita, H.~Shi, B.~Mutlu, and M.~Gleicher, ``{CollisionIK}: {A} {Per}-{Instant} {Pose} {Optimization} {Method} for {Generating} {Robot} {Motions} with {Environment} {Collision} {Avoidance},'' in \emph{2021 {International} {Conference} on {Robotics} and {Automation}}, May 2021, pp. 9995--10\,001.

\end{thebibliography}
\bibliographystyle{IEEEtran}

\end{document}